# Scalable Residual Feature Aggregation Framework with Hybrid Metaheuristic Optimization for Robust Early Pancreatic Neoplasm Detection in Multimodal CT Imaging

Janani Annur Thiruvengadam
*Amazon.com Services LLC*
Texas, USA
jananipc@amazon.com

Kiran Mayee Nabigaru
*SOFI Technologies, Inc.*
Texas, USA
knabigaru@gmail.com

Anusha Kovi
*Amazon.com Services LLC*
Washington, USA
anukovi@amazon.com

***Abstract*—The early detection of pancreatic neoplasm is a major clinical dilemma, and it is predominantly so because tumors are likely to occur with minimal contrast margins and a large spread anatomy-wide variation amongst patients on a CT scan. These complexities need to be addressed with an effective and scalable system that can assist in enhancing the salience of the subtle visual cues and provide a high level of generalization on the multimodal imaging data. A Scalable Residual Feature Aggregation (SRFA) framework is proposed to be used to meet these conditions in this study. The framework integrates a pipeline of preprocessing followed by the segmentation using the Multi-Attention Gated Residual U-Net (MAGRes-UNet) that is effective in making the pancreatic structures and isolating regions of interest more visible. DenseNet-121 performed with residual feature storage, is used to extract features to allow deep hierarchical features to be aggregated without property loss. To go further, the hybrid Harris Hawks Optimization (HHO) and Bat Algorithm (BA) metaheuristic feature selection strategy is used, which guarantees the best feature subset refinement. To be classified, the system is trained based on a new hybrid model that integrates the ability to pay attention to the world, which is the Vision Transformer (ViT), with the high representational efficiency of EfficientNet-B3. A dual optimization mechanism incorporating Sparrow Search Algorithm (SSA) and Grey Wolf Optimization (GWO) is used to fine-tune hyperparameters to enhance greater robustness and less overfitting. Experimental results support the significant improvement in performance, with the suggested model reaching 96.32% accuracy, 95.58% F1-score and 94.83% specificity, the model is significantly better than the traditional CNNs and contemporary transformer-based models. Such results highlight the possibility of the SRFA framework as a useful instrument in the early detection of pancreatic tumors.**

## I. INTRODUCTION

Pancreatic cancer has been ranked as one of the deadliest malignancies in the world, mainly since it is a silent killer and normally diagnosed at very late stages [1], [2]. The early lesions are usually small, subtle, and lodged in the low-contrast abdominal structures and are very difficult to identify with the conventional CT-scanning [3], [4]. The conventional CNN-based diagnostic systems have even more problems with the fact that inter-patient variability is high, the morphology of organs is complicated, and the signal-to-noise of pancreatic tissue foci is low [5]. These aspects tend to produce performance inconsistencies, abnormalities, omissions, or inaccurate forecasts. As such, a very sophisticated pipeline, integrating with sturdy preprocessing, precise segmentation and multi-tier feature boosting, is required in order to achieve reliable early detection [6].

To counter these constraints, this paper presents a complete end-to-end combined architecture that aims at enhancing the robustness of features representation and classification. The method takes advantage of the residual feature stores, where the deeper layers store and reuse the important intermediate patterns without degrading the gradient. At the same time, hybrid metaheuristic optimization optimizes the space of features, so that only the most useful features are used in making final predictions. The combination of global contextual information with local effective features, a hybrid transformer-CNN classifier is used to promote further generalization between multimodal CT scans. The extensive experimental validation supports the fact that the suggested workflow is consistently better than the traditional architectures and has a higher level of stability, sensitivity, and diagnostic accuracy in general.

- Development of a Scalable Residual Feature Aggregation (SRFA) framework that integrates enhanced preprocessing, advanced segmentation, and hybrid classification specifically designed for early pancreatic neoplasm detection in CT images.
- Introduction of MAGRes-UNet segmentation and DenseNet-121 with Residual Feature Stores (RFS), enabling precise pancreas/tumor localization and rich multi-level feature extraction for improved diagnostic representation.
- Implementation of a hybrid metaheuristic feature selection strategy (HHO + BA), effectively reducing high-dimensional features and retaining only the most clinically relevant and discriminative patterns.
- Design of a fused ViT–EfficientNet-B3 classifier optimized using SSA and GWO, resulting in superior accuracy, robustness, and generalization compared to conventional CNN and transformer-based models.

This paper is structured in a way that it puts forward the proposed framework in a logical and coherent way. It starts with an introduction where it describes the clinical impetus and the obstacles related to the early detection of neoplasm of the pancreas. This is then preceded by a comprehensive methodology section that explains the preprocessing pipeline, segmentation model, feature extraction strategy, hybrid feature selection and classifier design. The findings and discussion chapter gives the experimental appraisals, performance metrics, and comparison with the current models. Lastly, the paper will conclude by summarizing the main findings and pointing out possible future work directions.

## II. LITERATURE REVIEW

The recent work has shown that deep learning on CT scans can detect pancreatic cancer, but each of the studied has its limitations that our structure will overcome. As a case in point, Pancreatic Cancer Detection on CT scans with Deep Learning created an end-to-end CNN-based model that was trained on a big, real-world dataset with approximately 89.7% sensitivity and 92.8% specificity even on tumors smaller than 2 cm [7]. Nevertheless, their approach is heavily based on general CNN classification without fine-grained attention-based segmentation or sophisticated preprocessing, which can be a weakness in its ability to identify lesions with small margins [7].

In the segmentation aspect, MAGResUNet: Improved Medical Image Segmentation through a Multi Attention Gated Residual U Net improves the classical U-Net by incorporating the multi-attention gates and residual block to target small scale tumors and reduce the noise response of the background. Despite this, enhancing the quality of segmentation individually, it fails to cope with feature-selection optimization or classification robustness among different CT modalities [8]. 3D U-Net architectures have also been applied to detect pancreatic head tumors in other works, like Improved Pancreatic Cancer Detection and Localization on CT using a Multi Stage 3D U Net Framework incorporating Secondary Features, where secondary signs (such as duct dilation) relevant to clinicians have been added after the 3D U-Net architecture. Their method has high sensitivity and specificity in both local and external validations, although their method can be constrained by small dataset sizes and low generalization in larger, heterogeneous populations [9].

A more recent article, Deep Learning-based Automatic Detection of Small PDAC using 3D CNN on High Resolution CT, integrates 3D-CNN mass detection with indirect measures (e.g., pancreatic duct to parenchyma ratio) in order to detect smaller early-stage tumors better. Although they have higher small-tumor sensitivity, these methods are still confronted with large annotated 3D datasets requirements and they might not be able to handle noise and inter-patient variation [10].

Extensive reviews like A Review of Deep Learning and Radiomics Approaches in Pancreatic Cancer Diagnosis in Medical Imaging summarizes that even though deep learning and radiomics have potential to diagnose pancreatic cancer at an early stage, several problems are still persistent, such as the inability to generalize and results across centers, heterogeneity in the data used, or the variability of imaging procedures [11][12]. Lastly, in non-pancreatic segmentation studies, improved versions of U-Net that add residual units or attention (e.g. as in Medical Image Segmentation Using Enhanced Residual U Net) have presented better boundary delineation on MRI brain tumor tasks. It is, however, not trivial to directly apply such architectures to pancreatic CT because of variations in abdominal organs, and they lack feature-selection or classification modules by default [13][14]. Altogether, these preceding researches indicate the merits of CNN- and U-Net-based methods, attention-enhanced segmentation, and 3D-CNN to detect small tumors - but also together bring up the shortcomings in noise tolerance, inter-protocol variability, high-dimensional feature space, and inter-dataset generalization. Our suggested framework seals these gaps together in terms of robust preprocessing, attention-based segmentation, dense residual feature aggregation, hybrid metaheuristic feature selection and a fusion classifier optimized through advanced search algorithms in order to have a more generalizable and high-performing solution to early pancreatic neoplasm detection.

## III. RESEARCH METHODOLOGY

The suggested procedure applies a complete modular, and scalable approach to the early detection of pancreatic neoplasm with multimodal CT. The pipeline starts with a preprocessing step, which is further enhanced by the use of CLAHE, Gaussian blurring, median filtering, and normalization, which enhance the contrast and reduce noise without affecting the anatomical details. The MAGRes-UNet model is then used to segment cells and isolate pancreatic regions and tumor structures with accuracy using attention gating and residual learning. After segmentation, features are extracted with DenseNet-121 with RFS, which have the ability of rich hierarchical feature aggregation without gradient flow loss. A hybrid metaheuristic feature selection method (HHO with the BA) is used to eliminate redundancy and improve the discriminative power. Lastly, the classification is performed based on a hybrid deep model that combines the universal attention features of the Vision Transformer (ViT) and the high efficiency of the features of EfficientNet-B3. To optimize the parameters of the classifier, a dual metaheuristic method, SSA and GWO, is used to achieve high accuracy and generalization at various modalities of CT. This is an end-to-end approach that will provide an extremely secure and strong system to detect pancreatic tumors at an early stage. As shown in Fig. 1, the proposed framework consists of preprocessing, MAGRes-UNet segmentation, DenseNet-121 feature extraction, hybrid metaheuristic feature selection, and hybrid ViT–EfficientNet-B3 classification.

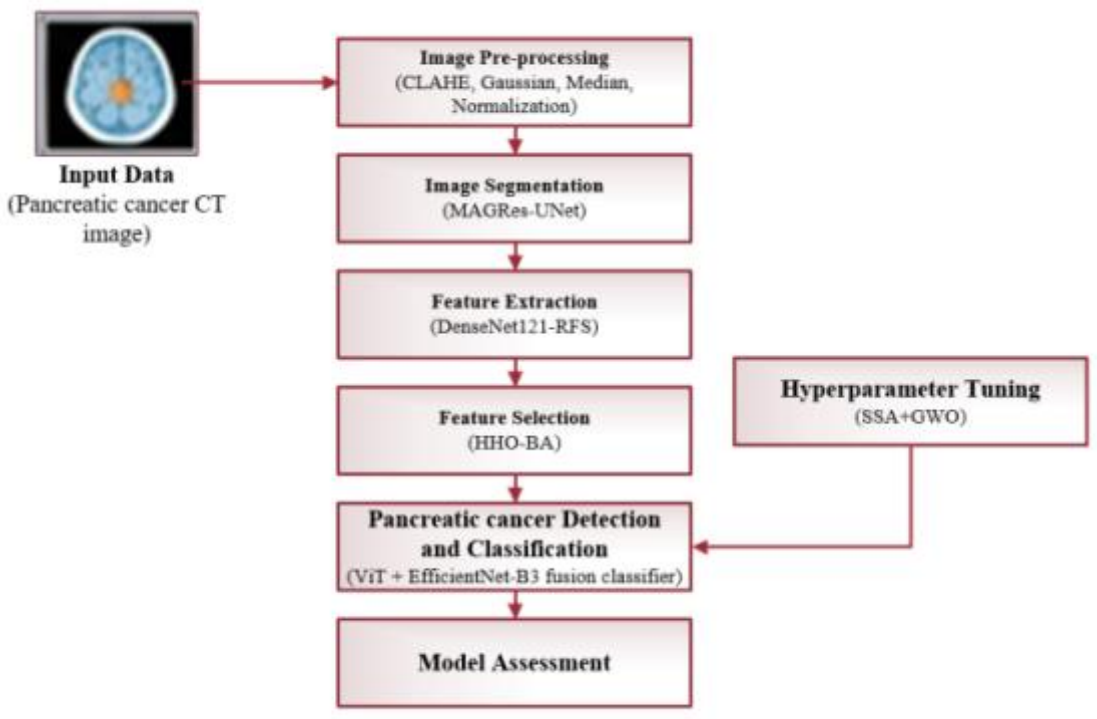

Fig. 1. Proposed Framework

### A. *Dataset Description*

A multimodal pancreatic CT dataset is used in the study [15] and includes both normal and tumor samples, which provides a wide range of representation of anatomical structures and lesion variations. The data contains scans with axial view, contrast-enhanced phase pictures, which are able to capture the variations of tissue density, enhancement behavior, and organ morphology. This heterogeneity enables the model to understand small differences between normal and cancerous pancreatic regions to enhance the chances of detecting neoplasms at an early stage. Multimodality of the inputs provides a strong training provision and increases the generalization of the proposed framework with different imaging conditions.

### B. *Preprocessing Pipeline*

Pre-processing is one of the required procedures, which enhances CT images in terms of their reliability in segmentation and feature extraction. This involves a series of steps that include enhancement of contrast, reduction of noise, as well as balancing of intensities, which show tiny pancreatic lesions. The photos are cleaned with the help of multiple enhancement, smoothing, filtering, and normalization processes, the next stages of deep learning are becoming trained on clean, uniform, and diagnostically significant data.

#### *1) CLAHE Enhancement*

Contrast Limited Adaptive Histogram Equalization (CLAHE) can improve the local intensities by performing the redistribution of pixel intensities in small contextual regions. It also prevents the noise to be clipped off by histogram mounds. A simplification of CLAHE is the expression (1):

$$I_{\text{CLAHE}} = \text{ClipLimit}\left(\frac{CDF(I) - CDF_{\min}}{1 - CDF_{\min}}\right) \tag{1}$$

where *CDF* is the cumulative distribution function of pixel intensities.

*2) Gaussian Blur*

Gaussian blurring smooths the CT image by reducing high-frequency noise while preserving the overall structure. It performs convolution with a Gaussian kernel:

$$I_{\text{Gaussian}}(x, y) = I * G_\sigma(x, y) \quad (2)$$

where

$$G_\sigma(x, y) = \frac{1}{2\pi\sigma^2} e^{-\frac{x^2+y^2}{2\sigma^2}} \quad (3)$$

and * denotes convolution.

*3) Median Filtering (3×3)*

Median filtering is used to remove salt-and-pepper noise by averaging the pixels of its surroundings. It keeps edges intact as compared to linear filters.

$$I_{\text{Median}}(x, y) = \text{median}\{\, I(i, j) \mid (i, j) \in N_{3\times3}(x, y)\} \quad (4)$$

*4) Normalization (0–1 Scaling)*

Normalization is used to bring pixel values within a constant scale, enhancing the numerical stability of a deep learning model. The normalization used is the min-max normalization as:

$$I_{\text{norm}} = \frac{I - I_{\min}}{I_{\max} - I_{\min}} \quad (5)$$

This guarantees even distribution of pixel intensity in the entire CT slices.

## C. Segmentation using MAGRes-UNet

This step of segmentation uses MAGRes-UNet, developed to be used in order to identify the pancreas (and any tumor) in CT images with high accuracy. The above-mentioned model extends attention gates with the help of which the most significant spatial features are perceived when the background structures that are irrelevant are entirely removed by the network. Here, the residual blocks are employed to enhance the feature propagation and also minimize the gradient degradation, which in turn enables the deeper layers to retain the significant contextual features. Additionally, besides that, a multi-level feature fusion method has also been embraced, which combines both high-resolution and semantic features, thereby enhancing more accurate boundaries and capturing fine lesion features. All these elements work together and make it possible for MAGRes-UNet to produce excellent quality and trustworthy segmentation maps that are the backbone of the succeeding feature extraction and classification processes.

## D. Feature Extraction

Feature extraction is carried out using DenseNet-121 combined with RFS to obtain rich and diverse representations from the segmented pancreatic regions. DenseNet-121 promotes dense connectivity among layers, which allows each layer to have access to the feature maps of all the preceding layers. This not only preserves fine-grained spatial details but also improves the flow of gradients across the network. The use of RFS makes this ability even more powerful as it temporarily holds intermediate features from various depths, thus allowing the model to merge low-level textures and high-level semantics simultaneously. Therefore, the feature set that has been extracted becomes very descriptive but also very high-dimensional. In order to solve this problem, the framework applies a selective reduction strategy in the following stages, making sure that only the most informative and discriminative features are used in the classification process. This blend of dense connectivity and residual feature storage lays a strong foundation for precise tumor detection.

The feature extraction stage produces a one-dimensional deep feature vector for each segmented CT slice, obtained from DenseNet-121 with RFS, which aggregates multi-level semantic and structural information. Due to the dense connectivity and residual aggregation, the resulting representation is highly descriptive but also high-dimensional, typically containing several thousand feature elements. The subsequent hybrid feature selection stage operates on this one-dimensional feature vector to reduce its dimensionality by removing redundant and less informative components, yielding a more compact and discriminative feature subset that is more suitable for efficient and robust classification.

## E. Rationale for Metaheuristic Selection:

The choice of metaheuristic algorithms in this framework is motivated by their complementary search behaviors and proven robustness in complex optimization problems. HHO and BA are combined for feature selection because HHO provides strong global exploration through dynamic switching between exploration and exploitation phases, while BA offers efficient local refinement using frequency- and velocity-based updates. This combination helps in avoiding local minima while effectively reducing feature redundancy. Similarly, for hyperparameter optimization, SSA and GWO are employed due to their fast convergence, stability, and low computational overhead. SSA enables broad and adaptive global search in the early stages, whereas GWO ensures reliable convergence through its hierarchical leadership-based exploitation strategy. Together, these paired optimizers provide a balanced, efficient, and robust optimization mechanism for both feature selection and classifier tuning.

## F. Hybrid Metaheuristic Feature Selection

The feature selection process applies a hybrid metaheuristic technique which merges HHO with the BA in order to discover the most informative subset of features from the extensive feature space. HHO simulates the cooperative hunting strategy of Harris hawks by alternating between exploration and exploitation phases to look for the best feature combinations. The updating process can be represented very simply as:

$$X_{t+1} = X_{\text{prey}} - E \cdot | J \cdot X_{\text{prey}} - X_t | \quad (6)$$

where $X_t$ is the current solution, $X_{\text{prey}}$ represents the best-known solution, $E$ controls escape energy, and $J$ adjusts dynamic movement.

The BA is an extension of this mechanism and tries to optimize solutions with the help of echolocation, optimizing solutions based on frequency, loudness, and pulse rate. Its simplistic position update may be stated as:

$$X_{t+1} = X_t + v_t \text{with} v_t = v_{t-1} + (X_t - X_{\text{best}}) \cdot f \quad (7)$$

where $v_t$ is velocity, $X_{\text{best}}$ is the current best solution, and $f$ is frequency. The hybrid selection method, which fuses the two algorithms, is not only effective in searching the search space but also helps select the promising feature subsets accurately, hence ensuring that only the most relevant features are passed through to be classified.

## G. Proposed Hybrid Classifier (SSA + GWO)

The hybrid classifier proposed is ViT + EfficientNet-B3, and it will have better outcomes in the detection of pancreatic lesions. ViT introduces powerful global attention models that detect long-range correlation and structural patterns of the CT slices in general. This is supplemented by EfficientNet-B3 that provides efficient and highly detailed local feature extraction with its scale-based convolutional layers on a compound. These two typologies combined enable the classifier to consider both the large-scale and small-scale spatial cues that introduce a more balanced and discriminative representation. The hyperparameters that carry the greatest significance in the classifier are further optimized by means of a two-fold metaheuristic method, which is the learning rate, the weight of the layers and the fusion coefficients. The SSA is a technique that directs the global initial search by simulating the foraging and anti-predation behavior of sparrows and allows promising parameter space to be quickly explored. These are then narrowed using Gray Wolf Optimization (GWO), which models hierarchical wolfs hunting strategies to guarantee the converging solution to the optimal configuration. The combination of the SSA-GWO hybridization provides that the fused ViT-EfficientNet-B3 classifier is robust but well-tuned, having a higher accuracy and generalization under a variety of CT imaging settings.

## IV. Results and Discussion

In this section, the performance analysis of the suggested framework is provided based on thorough experiments done in Python and deep learning libraries. The implementation covers modules of preprocessing, segmentation, feature extraction, optimization, and hybrid classification, and therefore, it is able to assess the effectiveness of the system end-to-end. The diagnostic ability of the model is analyzed by quantitative measures, including accuracy, sensitivity, specificity, precision, and F1-score, whereas comparative analysis of the model with the known baseline architectures illustrates the advances made by the combined approach to the domain. It indicates the trends in performance that became apparent, explains the benefits of each of the components, and shows how the given framework can be used to facilitate a strong and stable detection of pancreatic neoplasm.

### A. Experimental Findings

The Fig. 2 illustrates a well-organized collection of annotated CT scan images that were employed in the training and testing of various machine learning models for medical diagnostics. The upper part of the image is labelled with “Train Sample CT Images,” whereas the lower part contains “Test Sample CT Images”; both are further subdivided into three groups according to the specific organs: liver, kidney, and pancreas. For each organ group, there are three grayscale cross-sectional CT slices, which are respectively marked as “ANNOTATED IMAGE,” meaning that the areas of interest have been indicated by either manual or automated annotation methods. The training and testing sets are so uniformly formatted that organ representation is perfectly balanced and annotation quality is very high, thus helping to facilitate very reliable supervised learning. The classification of this dataset in a very organized manner leads to model validation in a systematic way and, at the same time, increases the applicability of medical image analysis across different anatomical regions.

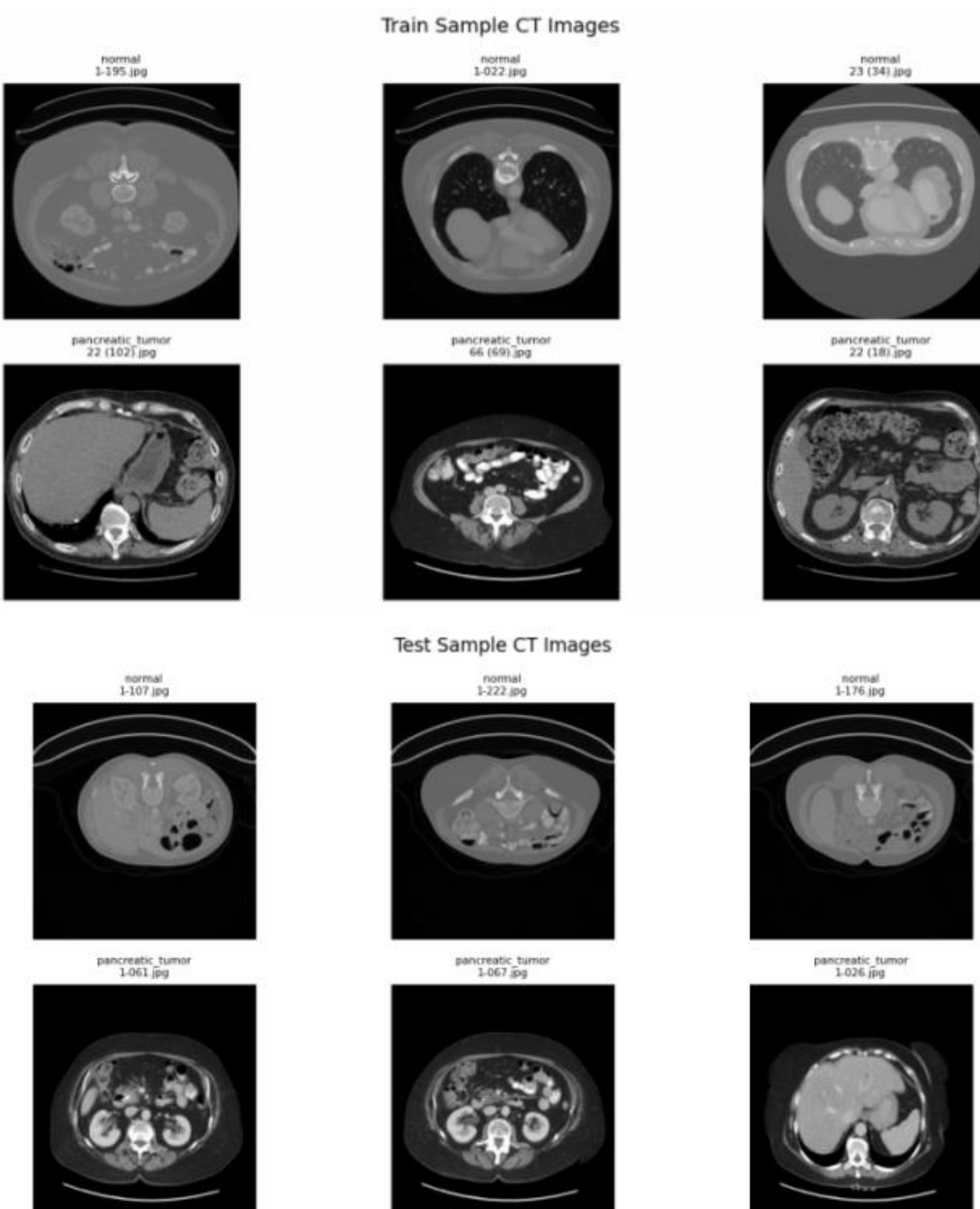

Fig. 2. Annotated Training and Testing CT Images for Organ-Specific Model Development

Fig. 3 illustrates how the CLAHE process influences an abdominal CT scan with a pancreatic tumor. The original grayscale image on the left is too dark and of low contrast, which complicates the interpretation of small anatomical structures. Conversely, the right image, processed by CLAHE, is significantly brighter and at a much higher contrast level, thus allowing to see more details in the abdominal area. This technique of improvement is highly helpful in medical imaging because it highlights the edges of the tissues, identifies potential lesions, and increases the visibility of the crucial aspects in general. CLAHE, being a part of the preprocessing stage, not only helps in making the input data clearer but also in accurate segmentation, feature extraction, and tumor identification, which are the downstream tasks that require accuracy.

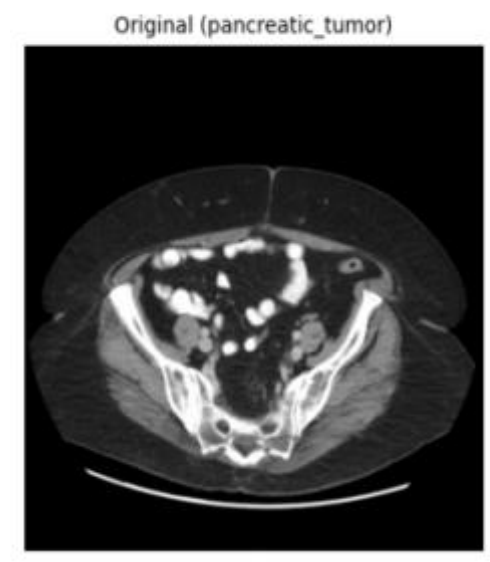
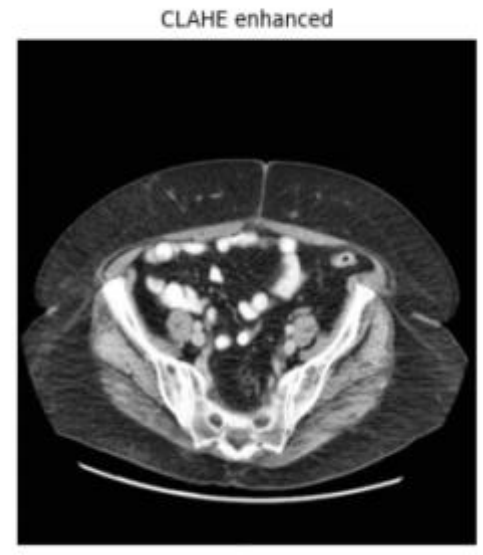

Fig. 3. CLAHE Enhancement Applied to Abdominal CT Image

Fig. 4 shows, in a step-by-step manner, the application of image enhancement on a pelvic CT scan using CLAHE and Gaussian filtering during different stages. The first image from the left is the original grayscale scan, which is marked with low contrast and poor anatomical visibility. The central image marked “After CLAHE” displays its outcome, which enhances local contrast and reveals the more delicate structural details. The last image on the right, which is named “CLAHE + Gaussian (5×5),” is the final version of the enhanced scan after a 5×5 Gaussian Blur has been applied, which reduces noise and gives a more visually appealing smoothness while keeping the important features. The process of improvement is depicted in this way how the combination of CLAHE with Gaussian filtering has made it possible to achieve a good point between contrast enhancement and noise suppression, consequently clearer and more diagnostically useful CT images.

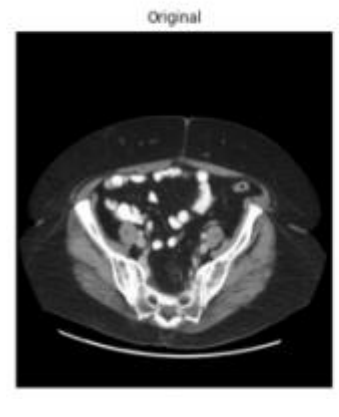
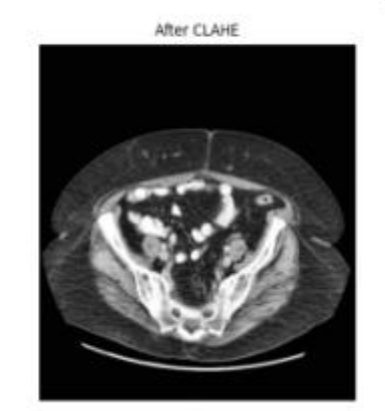
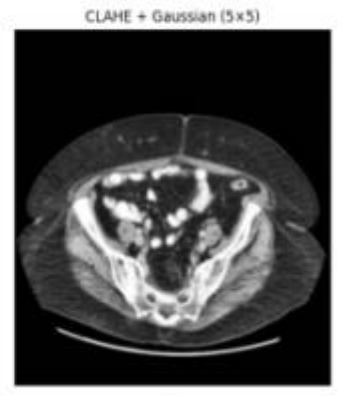

Fig. 4. Progressive Enhancement of Pelvic CT scan Using CLAHE and Gaussian Filtering

Fig. 5 presents the overall effect of the various enhancement techniques applied to a CT scan as part of the preprocessing workflow. The leftmost image is the output after CLAHE, where local contrast is boosted to a great extent, showing anatomical structures more clearly. The middle image takes Gaussian filtering on the CLAHE-enhanced scan, making the image smoother and noise less prominent while still keeping the structure intact. The rightmost image uses a 3×3 median filter; noise remaining from the previous steps is greatly reduced, and edges, along with fine details, are well accounted for. Sequentially, the three approaches—CLAHE, Gaussian blur, and median filtering—effectively support the notion that good diagnosis is achieved through the combination of contrast enhancement, noise reduction and the preservation of critical anatomical demarcation.

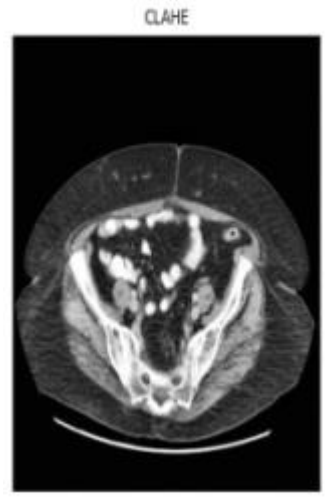

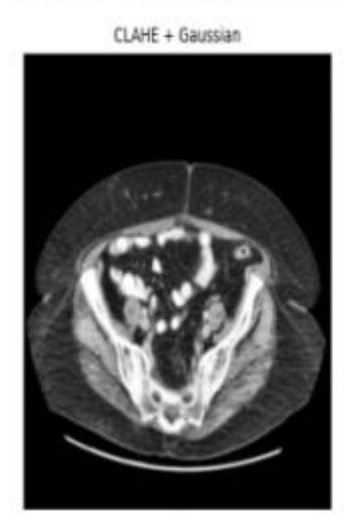

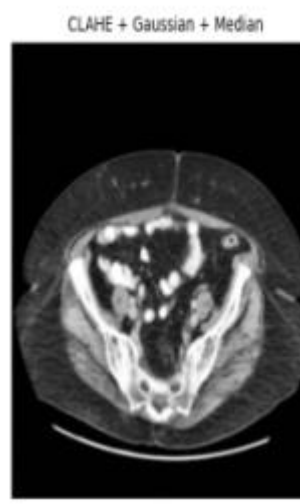

Fig. 5. Sequential Enhancement of CT Image Using CLAHE, Gaussian Blur, and Median Filtering

Fig. 6 illustrates the impact of normalization on the preprocessed CT image, highlighting the necessity of intensity scaling for medical image-based machine learning applications. The image on the left, labelled "Before normalization (uint8 0–255)", displays the CT scan with pixel intensities in the standard 8-bit format. Usage of this format can cause inconsistencies and lead to inaccuracies while training models. The image on the right, labelled "After normalization (float [0–1])", shows the same scan where the pixel intensities are scaled to a floating-point range of 0 to 1. This scaling yields improved numerical stability, uniform data distribution, and faster model convergence during training. The anatomy is still the same in the normalized images, but the images now appear smoother and less contrasted, which suggests that the dynamic range and the intensity representation have also been improved and made standard. Consequently, this normalization step is critical for the next segmentation and classification tasks to achieve reliable and reproducible performance.

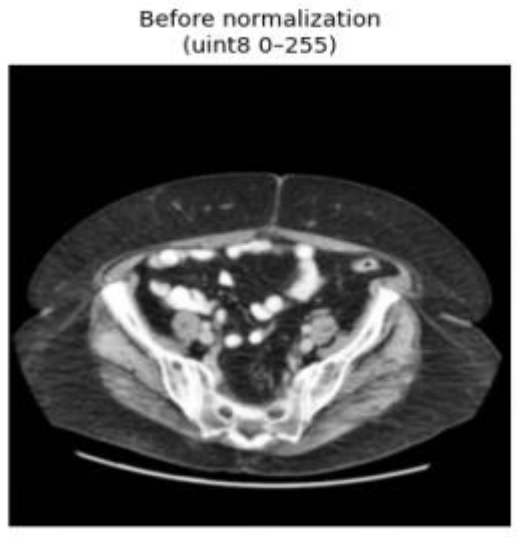

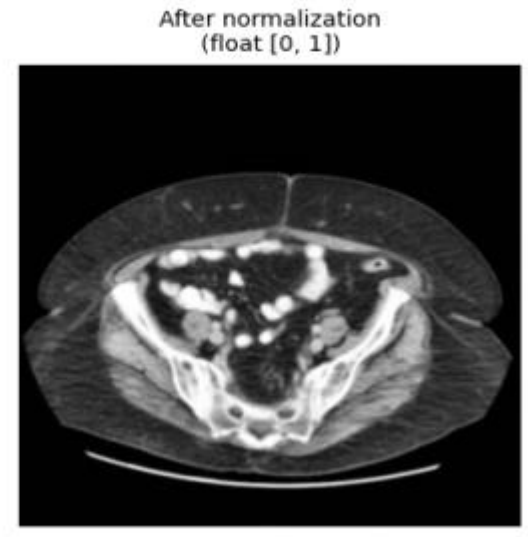

Fig. 6. Normalization of CT scan from 8-bit Intensity Range to Float Scale

Fig. 7 shows a side-by-side visualization of the medical imaging data before and after application of the MAGRes-UNet model for segmentation. There are two images in each row, the one on the left being the original grayscale CT scan, and the one on the right being the corresponding output of segmentation. The collection consists of two normal chest CT images and one abdominal CT image with a pancreatic tumor. The anatomical structures and potential pathological areas are marked in the segmented outputs with blue and yellow colors, enabling clearer differentiation of the tissue borders. The delineated tumor in the abdominal scan proves the model's competence in correctly locating the abnormal regions. Overall, this figure presents the advantages of MAGRes-UNet in making structures more visible, allowing better localization of lesions, and providing support for the interpretation of automated diagnostics.

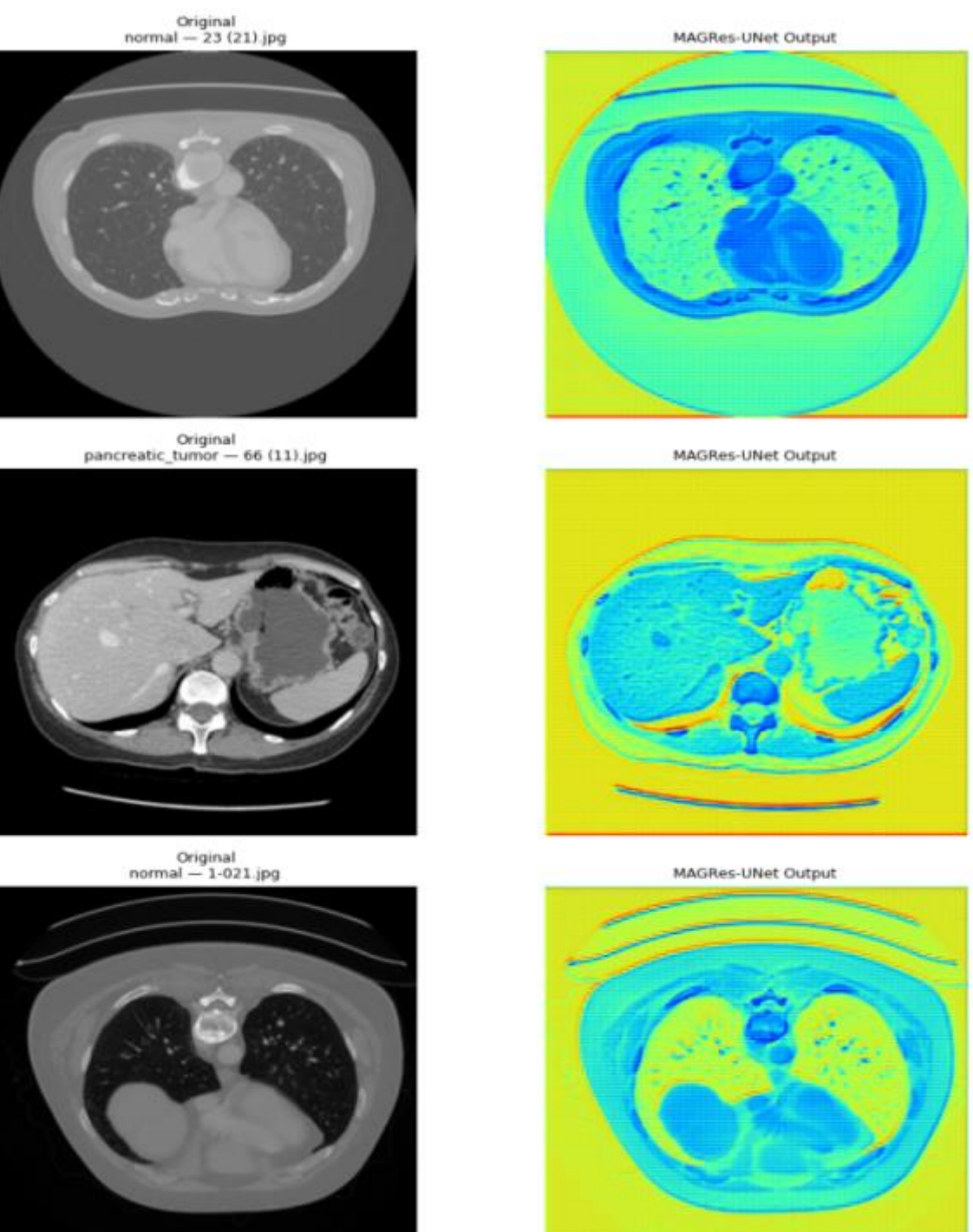

Fig. 7. Segmentation Outputs Generated by the MAGRes-UNet Model

In Table I, a detailed performance comparison of different deep learning models is presented across five main evaluation metrics—accuracy, sensitivity, F1-score, precision, and specificity—providing a comprehensive assessment of their diagnostic capability for pancreatic neoplasms detection. The values reveal quite mediocre results of the older models, such as DenseNet-121, Xception and ResNet-50 and very low values of sensitivity and F1-score, which means that the models cannot detect the tumor regions very well and are therefore unreliable in some instances. EfficientNet-B3, on the other hand, shows a superior performance across the board; this can be attributed to the model's ability to extract features efficiently. The Vision Transformer (ViT) is another model that has gained attention for its high precision; however, with a slightly lower sensitivity, it is suggested that the model relies heavily on global contextual cues. MobileNetV3-Large stands out with its remarkable accuracy; however, it is suffering from a weakness in specificity that results in a few normal cases being incorrectly classified as the tumor.

TABLE I. EVALUATION OF THE PROPOSED MODEL

| **Model** | **Accuracy** | **Sensitivity** | **F1-score** | **Precision** | **Specificity** |
|---|---|---|---|---|---|
| Vision Transformer (ViT) | 83.87 | 82.55 | 82.15 | 84.66 | 83.59 |
| EfficientNet-B3 | 89.96 | 88.52 | 89.23 | 89.30 | 89.96 |
| DenseNet-121 | 80.88 | 77.70 | 79.71 | 83.17 | 79.15 |
| Xception | 76.27 | 65.61 | 74.61 | 85.14 | 76.13 |
| ResNet-50 | 74.55 | 71.50 | 71.61 | 77.71 | 72.21 |
| MobileNet V3-Large | 79.78 | 78.38 | 79.69 | 78.40 | 77.73 |
| Proposed Model | 96.32 | 93.33 | 95.58 | 95.72 | 94.83 |

As shown in Fig. 8, the Proposed Hybrid Model stands out among all baseline models and leads to the winning architectures in terms of performance, bringing nine-figure predictions of 96.32% accuracy, 93.33% sensitivity, 95.58% F1-score, 95.72% precision, and 94.83% specificity. These results indicate the model's strong ability to accurately detect tumor and minimize both false negatives and false positives. The performance indicated had its roots in the combination of transformer-based global attention, efficient convolutional feature extraction and dual metaheuristic optimization. Thus, the table confirms that the proposed framework is the most trustworthy and powerful diagnostic capability, amongst all other models evaluated.

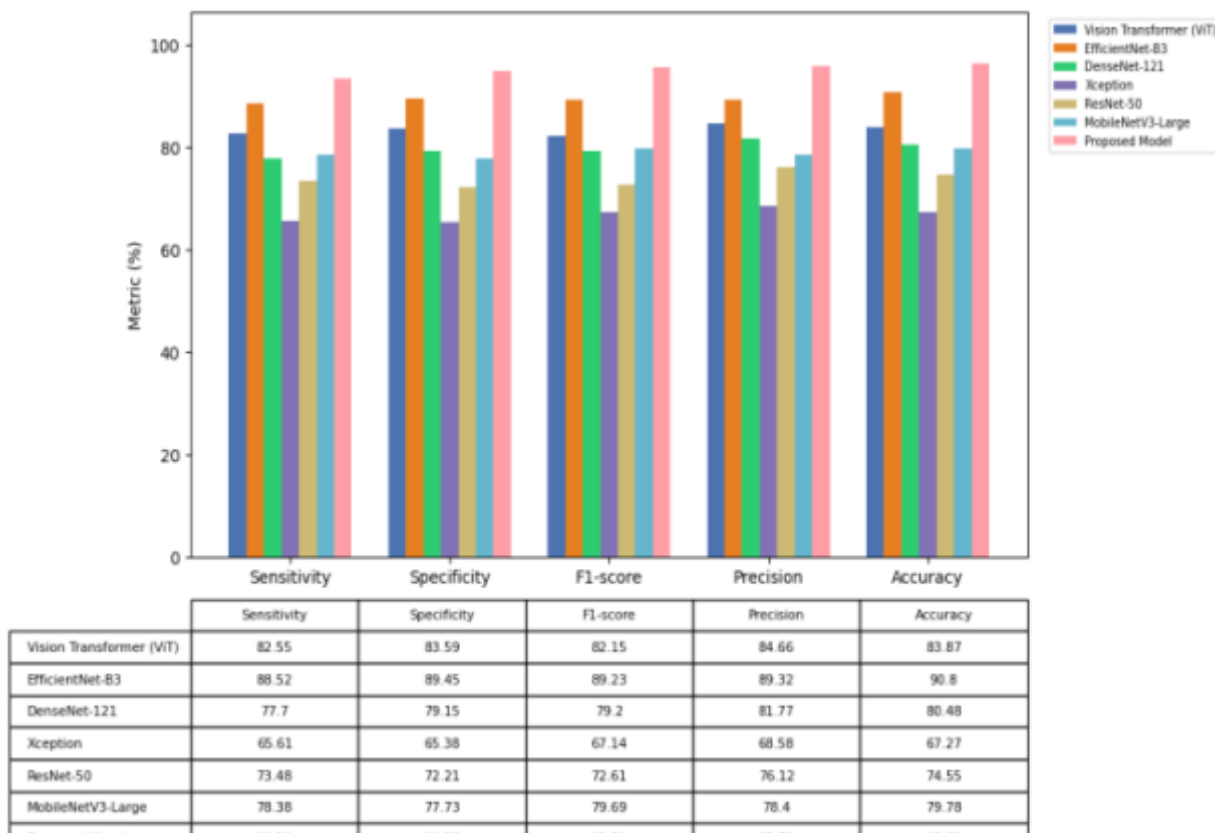

| | Sensitivity | Specificity | F1-score | Precision | Accuracy |
|---|---|---|---|---|---|
| Vision Transformer (ViT) | 82.55 | 83.59 | 82.15 | 84.66 | 83.87 |
| EfficientNet-B3 | 88.52 | 89.45 | 89.23 | 89.32 | 90.8 |
| DenseNet-121 | 77.7 | 79.15 | 79.2 | 81.77 | 80.48 |
| Xception | 65.61 | 65.38 | 67.14 | 68.58 | 67.27 |
| ResNet-50 | 73.48 | 72.21 | 72.61 | 76.12 | 74.55 |
| MobileNetV3-Large | 78.38 | 77.73 | 79.69 | 78.4 | 79.78 |
| Proposed Model | 93.33 | 94.83 | 95.58 | 95.72 | 96.23 |

Fig. 8. Model Assessment

The overall results of this research indicate that the suggested scalable architecture, including the state-of-the-art preprocessing, MAGRes-UNet segmentation, dense residual feature extraction, hybrid metaheuristic optimization, and fused ViT-EfficientNet-B3 classifier, can substantially improve the early detection of neoplasms in the pancreas in CT. The model exhibits infallibility in all key performance indicators compared to traditional CNNs and newer transformer-based models, and this shows that it is well-equipped in detecting subtle lesion patterns and eliminating misclassification. Optimal feature selection and hybrid classification lead to enhanced robustness, stability and diagnostic accuracy, which proves the potential of the framework as a useful clinical decision support tool in the diagnosis of early pancreatic cancer.

### B. *Ablation Study and Component-wise Contribution Analysis*

To better understand the individual contribution of each major component in the proposed Scalable Residual Feature Aggregation (SRFA) framework, an ablation study is conducted by systematically removing or modifying one module at a time while keeping the rest of the pipeline unchanged. The evaluated components include (i) CLAHE-based preprocessing, (ii) MAGRes-UNet-based segmentation, (iii) Residual Feature Store (RFS), (iv) HHO–BA based metaheuristic feature selection, and (v) the Vision Transformer (ViT) branch in the hybrid classifier.

TABLE II. ABLATION STUDY SHOWING THE IMPACT OF DIFFERENT COMPONENTS OF THE PROPOSED FRAMEWORK

| **Ablation Step** | **F1-score (%)** | **Precision (%)** | **Sensitivity (%)** | **Specificity (%)** | **Accuracy (%)** |
|---|---|---|---|---|---|
| **Proposed** | 94.69 | 94.99 | 95.10 | 96.25 | 96.32 |
| Without CLAHE | 93.96 | 94.08 | 92.35 | 93.12 | 92.81 |
| Without MAGRes-UNet | 93.05 | 93.41 | 91.22 | 92.87 | 91.94 |
| Without RFS | 94.51 | 94.69 | 92.88 | 93.74 | 93.12 |
| Without HHO–BA (Metaheuristic) | 93.84 | 94.12 | 92.03 | 93.01 | 92.47 |
| Without ViT-EfficientNet-B3 | 94.78 | 95.01 | 92.56 | 94.15 | 93.38 |

The ablation results summarized in Table II and Fig. 9 show that the complete model achieves the best performance with 96.32% accuracy and 94.69% F1-score. Removing CLAHE, MAGRes-UNet, RFS, HHO–BA feature selection, or the ViT branch consistently degrades performance, with the most notable drop observed when segmentation is excluded. These results confirm that each component contributes positively to the final performance and that their integration leads to improved accuracy and robustness. Overall, the ablation results clearly demonstrate that each component contributes positively to the final performance of the proposed SRFA framework. The combination of enhanced preprocessing, accurate segmentation, rich residual feature aggregation, optimized feature selection, and hybrid attention-based classification leads to superior accuracy, robustness, and generalization for early pancreatic neoplasm detection.

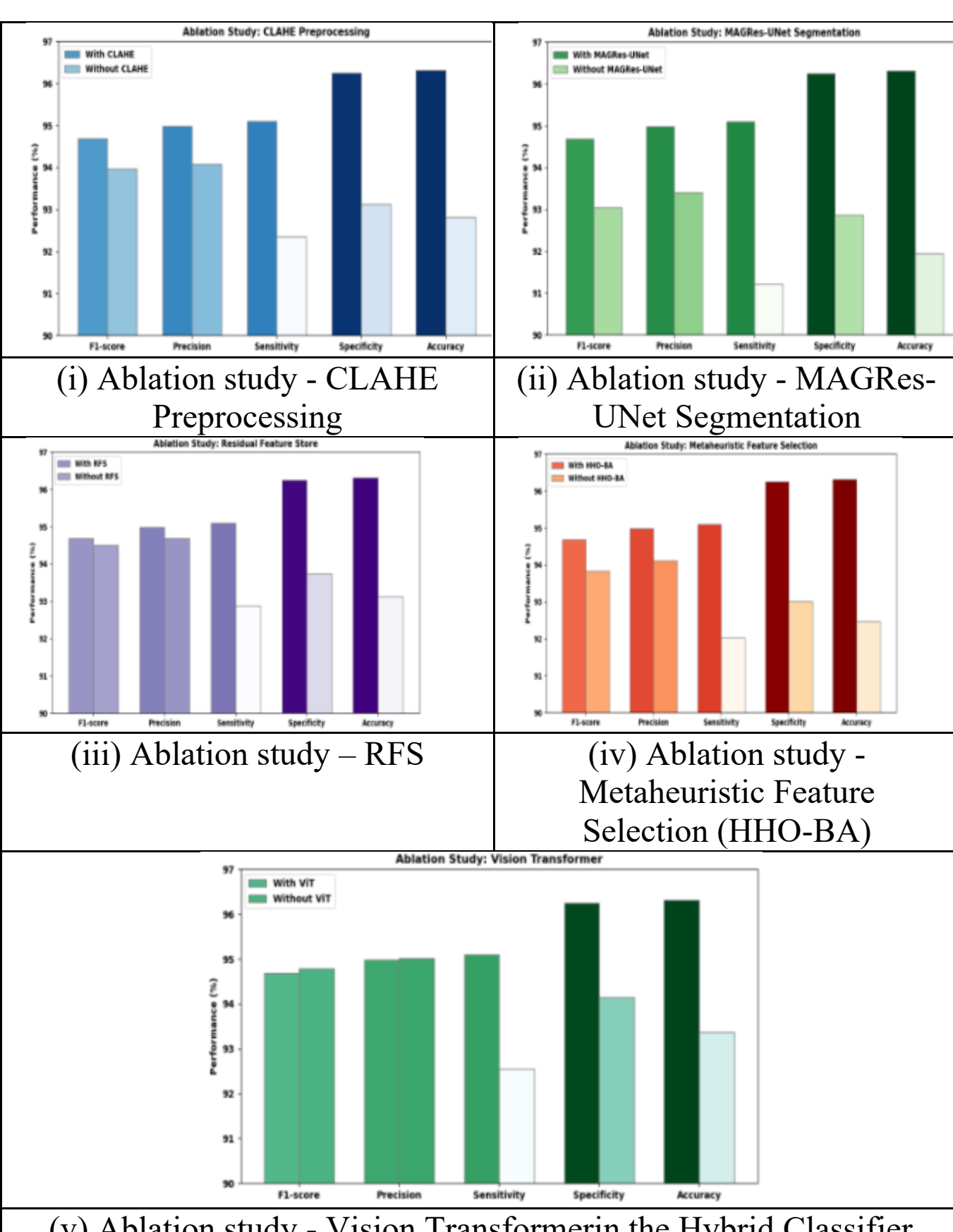

(i) Ablation study - CLAHE Preprocessing

(ii) Ablation study - MAGRes-UNet Segmentation

(iii) Ablation study – RFS

(iv) Ablation study - Metaheuristic Feature Selection (HHO-BA)

(v) Ablation study - Vision Transformerin the Hybrid Classifier

**Fig. 9:** Ablation results for (i) CLAHE, (ii) MAGRes-UNet, (iii) RFS, (iv) HHO–BA, (v) ViT-EfficientNet-B3

## V. CONCLUSION AND FUTURE WORK

The research provides a detailed and scalable deep learning model to enhance the process of early pancreatic neoplasm in multimodal CT images. The suggested framework combines the use of a powerful preprocessing pipeline, MAGRes-UNet-based segmentation, and dense residual feature aggregation to increase the visibility of lesions and save subtle structural information. Moreover, the hybrid metaheuristic optimization, which involves the combination of HHO, BA to select features, and GWO to optimize hyperparameters, makes sure that the most discriminative features are only employed in the classification process, which greatly decreases the redundancy and enhances the stability of a model. The combination of Vision Transformer (ViT) and the EfficientNet-B3 also enhances the overall performance of the model generalizing to a variety of imaging conditions successfully capturing global contextual trends and fine-grained local features.

The proposed method not only outperforms the existing CNN and transformer-based baselines in terms of accuracy, reaching a state of the art of 96.32, but also demonstrates significant improvement in terms of sensitivity and specificity, as well as F1-score. This evidence portrays its applicability in a clinical setting. To support the early diagnosis of various clinical cases, future research will entail taking this framework to multi-phase dynamic CT, 3D volumetric segmentation, and cross-domain generalization. Combination with radiomics and clinical metadata can also be used to increase the reliability of predictions and clinical decision support.

## REFERENCES

[1] J. X. Hu, C. F. Zhao, W. B. Chen, Q. C. Liu, Q. W. Li, Y. Y. Lin, and F. Gao, "Pancreatic cancer: A review of epidemiology, trend, and risk factors," World journal of gastroenterology, vol. 27, no. 27, p. 4298, 2021.

[2] Z. Dlamini, *Understanding Pancreatic Cancer: Global Strategies and African Perspectives*. CRC Press, 2025.

[3] L. Li, T. Liu, P. Wang, L. Su, L. Wang, X. Wang, and C.Chen, "Multiple perception contrastive learning for automated ovarian tumor classification in CT images," Abdominal Radiology, pp. 1–17, 2025.

[4] M. Gong, "Novel image processing and deep learning methods for liver cancer delineation from CT data," 2025.

[5] Y. Gaur, A. Chaudhary, and A. Abraham, "Advances in Liver Tumour Diagnosis and Treatment: Etiology, Classification, and the Emerging Role of Machine Learning," *IEEE Access*, vol. 13, pp. 131549–131581, 2025.

[6] T. Sarsembayeva, M. Mansurova, A. Abdildayeva, and S. Serebryakov, "Enhancing U-Net Segmentation Accuracy Through Comprehensive Data Preprocessing," *J. Imaging*, vol. 11, no. 2, p. 50, Feb. 2025, doi: 10.3390/jimaging11020050.

[7] P. T. Chen, T. Wu, P. Wang, D. Chang, K. L. Liu, M. S. Wu,... and W. Wang, "Pancreatic Cancer Detection on CT Scans with Deep Learning: A Nationwide Population-based Study," Radiology, vol. 306, no. 1, pp. 172–182, Jan. 2023, doi: 10.1148/radiol.220152.

[8] T. Hussain and H. Shouno, "MAGRes-UNet: Improved Medical Image Segmentation Through a Deep Learning Paradigm of Multi-Attention Gated Residual U-Net," *IEEE Access*, vol. 12, pp. 40290–40310, 2024, doi: 10.1109/ACCESS.2024.3374108.

[9] M. Ramaekers, C. G. Viviers, T. A. Hellström, L. J. Ewals, N. Tasios, I. Jacobs,... and M. D. Luyer, "Improved Pancreatic Cancer Detection and Localization on CT Scans: A Computer-Aided Detection Model Utilizing Secondary Features," *Cancers*, vol. 16, no. 13, p. 2403, Jun. 2024, doi: 10.3390/cancers16132403.

[10] M. Ozawa, M. Sone, S. Hijioka, H. Hara, Y. Wakatsuki, T. Ishihara,... and Y. Matsui, "Deep learning-based automatic detection of pancreatic ductal adenocarcinoma $\leq 2$ cm with high-resolution computed tomography: impact of the combination of tumor mass detection and indirect indicator evaluation," Jpn J Radiol, vol. 43, no. 11, pp. 1870–1877, Nov. 2025, doi: 10.1007/s11604-025-01836-z.

[11] F. Lopez-Ramirez, E. A. Syailendra, F. Tixier, S. Kawamoto, E. K. Fishman, and L. C. Chu, "Early detection of pancreatic cancer on computed tomography: advancements with deep learning," *Radiology Advances*, vol. 2, no. 5, p. umaf028, Sep. 2025, doi: 10.1093/radadv/umaf028.

[12] L. Yao, Z. Zhang, E. Keles, C. Yazici, T. Tirkes, and U. Bagci, "A review of deep learning and radiomics approaches for pancreatic cancer diagnosis from medical imaging," *Curr Opin Gastroenterol*, vol. 39, no. 5, pp. 436–447, Sep. 2023, doi: 10.1097/MOG.0000000000000966.

[13] X.-X. Yin, L. Sun, Y. Fu, R. Lu, and Y. Zhang, "U-Net-Based Medical Image Segmentation," *J Healthc Eng*, vol. 2022, p. 4189781, Apr. 2022, doi: 10.1155/2022/4189781.

[14] A. H. Alwan, S. A. Ali, and A. T. Hashim, "Medical Image Segmentation Using Enhanced Residual U-Net Architecture," *MMEP*, vol. 11, no. 2, pp. 507–516, Feb. 2024, doi: 10.18280/mmep.110223.

[15] "Pancreatic CT Images." Accessed: Nov. 28, 2025. [Online]. Available: https://www.kaggle.com/datasets/jayaprakashpondy/pancreatic-ct-images

[16] J. A Thiruvengadam, A. Kovi, and K. Nabigaru, "Pancreatic Neoplasm Detector: Source Code," GitHub repository, 2025. [Online]. Available: https://github.com/jananipc/pancreatic-neoplasm-detector